%% file: main.tex
\ifdefined\pdfminorversion
\fi
\documentclass{article}
\usepackage{microtype}
\usepackage{graphicx}
\usepackage{booktabs}
\usepackage{hyperref}
\usepackage[preprint]{icml2027}
\usepackage{amsmath,amssymb,mathtools,amsthm}
\newtheorem{lemma}{Lemma}
\newtheorem{corollary}{Corollary}
\newtheorem{proposition}{Proposition}
\theoremstyle{definition}
\newtheorem{assumption}{Assumption}
\theoremstyle{remark}
\newtheorem{remark}{Remark}
\icmltitlerunning{Stop Committing: A Verification View of Multi-Image Inference}
\newcommand{\RoneIV}{.787}
\newcommand{\RoneQV}{.832}
\newcommand{\CtopFive}{.95}
\newcommand{\AccOracle}{.901}
\newcommand{\AccGAVRIV}{.834}
\newcommand{\GainNfifty}{18.4--20.5}
\newcommand{\GainAvg}{5.1}
\newcommand{\RecoveredPct}{44.7\%} 
\providecommand{\McqGainMax}{3.0}

\begin{document}
\twocolumn[
  \icmltitle{Stop Committing: A Verification View of Multi-Image Inference}
  \begin{icmlauthorlist}
    \icmlauthor{Zhirui Chen}{mpu}
    \icmlauthor{Haiquan Liu}{mpu}
    \icmlauthor{Haiyang Yu}{mpu}
    \icmlauthor{Jieru Zhang}{mpu}
    \icmlauthor{Tao Tan}{mpu}
    \icmlauthor{Ling Shao}{ucas}
  \end{icmlauthorlist}
  \icmlaffiliation{mpu}{Macao Polytechnic University, Macao, China}
  \icmlaffiliation{ucas}{University of Chinese Academy of Sciences, Beijing, China}
  \icmlcorrespondingauthor{Tao Tan}{taotan@mpu.edu.mo}
  \icmlcorrespondingauthor{Ling Shao}{ling.shao@ieee.org}
  \icmlkeywords{Multi-Image Inference, Vision--Language Models, Visual Question Answering}
  \vskip 0.3in
]
\printAffiliationsAndNotice{}
\begin{abstract}
Answering an image-grounded question in a multi-image context requires
identifying the image the question refers to. Models often select the
image implicitly and answer in the same pass, leaving no opportunity
to recover from a selection error. We decompose end-to-end accuracy
into candidate coverage, verification error, and readout accuracy.
Guided by this analysis, we introduce \textbf{GAVR}, a training-free
wrapper that retains the top-$k$ images ranked by the model's internal
pointer, asks the model whether each candidate contains the question's
anchor, and answers the original question using the verified image
alone. The verification score ranks candidates by the model's
affirmative probability, while single-image readout avoids dilution
by distractors. Our analysis predicts larger gains as the haystack
grows and smaller gains when initial retrieval is already strong. Each verification and readout
pass processes one image, using the same frozen model.
On Visual Haystacks, GAVR improves over joint inference by
\GainNfifty{} points at $N{=}50$ and over pointer top-1 by $\GainAvg$
points on average at $N{=}10$ across five models. On a
6{,}000-example multi-image multiple-choice benchmark, gains are at
most $\McqGainMax$ points. Our source code is available at
\url{https://github.com/Zhirui-Chen-2/GAVR}.
\end{abstract}

\section{Introduction}
\label{sec:intro}

\begin{figure}[t]
  \centering
  \includegraphics[width=\columnwidth]{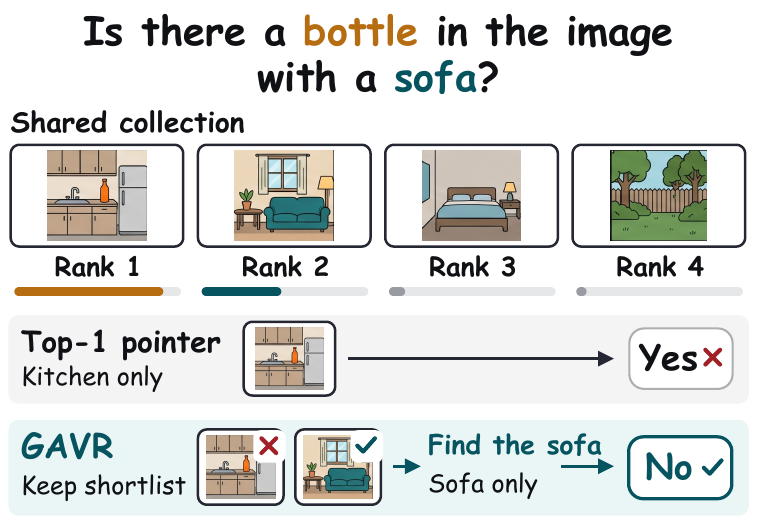}
  \caption{A top-1 pointer selects a kitchen with a bottle. GAVR checks
  the shortlist for a sofa and answers from the verified living room,
  where there is no bottle.}
  \label{fig:gavr-teaser}
\end{figure}

A question about several images can use one object to identify the
relevant image and ask about a different object in it. To decide
whether there is a bottle in the photo with the sofa, a model must
find the sofa before checking for a bottle. Figure~\ref{fig:gavr-teaser}
shows how selecting the wrong image can produce a false \emph{yes}
even when the sofa image is in the shortlist. As the image pool grows,
distractors can interfere with both decisions. Visual Haystacks
isolates this problem in image collections, while MultiHaystack
extends evidence retrieval and reasoning to collections of images,
videos, and documents~\citep{wu2025visual,xu2026multihaystack}.

Recent work exposes retrieval signals inside vision-language models.
Attention over visual tokens can rank images or document regions,
and head-level analyses trace how information about a target image
or region reaches later layers
~\citep{wu2025retrieval,park2026irh,li2026multimodalheads,park2026vrh}.
These signals can supply a shortlist for answering. A pipeline that
uses only the first-ranked image discards plausible evidence before
checking the referring condition in the question. In our Visual
Haystacks experiments with ten images, the target appears among the
first five candidates in about 95\% of examples on average. A wrong
first choice can therefore leave the right image available for
verification.

We introduce \textbf{GAVR}, short for Grounded Anchor Verification
and Re-conditioning, a training-free inference procedure for a frozen
vision-language model. \emph{Candidate retention} keeps the top-$k$
ranked images so a target below rank one remains eligible.
\emph{Grounded anchor verification} asks the model whether each
candidate contains the anchor named in the question. In \emph{focused
re-conditioning}, the same model answers the original question with
the highest-scoring image as its only visual input.

GAVR uses the pointer score studied by \citet{park2026irh} to rank
candidates. Related inference methods use retrieval-head attention
to verbalize visual evidence or use language-generated hypotheses
to rerank multimodal search results~\citep{pei2026vera,abdalla2026hive}.
Because GAVR makes image selection explicit, we can distinguish
whether the candidate set contains the target, whether the verifier
selects it, and whether the model answers correctly from it.

On Visual Haystacks with ten images, GAVR improves answer accuracy
by 1.5 to 6.6 percentage points across five models over top-1 pointer
selection followed by single-image answering at the same resolution.
At 50 images, it improves accuracy by 18.4 to 20.5 points across four
models over joint multi-image answering. Gains are smaller on the
6{,}000-example ICIR-MCQ evaluation, where each question has five
or six candidate images and initial retrieval is already strong.
Together, these results show where verification has room to help:
the ranking can miss at rank one while still retaining the target
in a small set that the model can check.

\section{Related Work}
\label{sec:related}

\paragraph{Locating evidence in a shared visual context.}
Visual Haystacks varies the number of distractors around a target
image~\citep{wu2025visual}, while MultiHaystack tests evidence retrieval
across images, videos, and documents~\citep{xu2026multihaystack}.
MMRA asks about relations across images~\citep{wu2026mmra}, and
Distract-Bench tests whether models reason from irrelevant visual
cues even when the answer is fixed~\citep{sun2026distractbench}.
These settings call for different evidence budgets. We focus on
questions whose referring phrase identifies one image, so a model
can verify that reference independently of the requested attribute
and answer from the selected source alone.

\paragraph{From internal retrieval to candidate coverage.}
Retrieval heads have been studied as mechanisms for locating
information in long text contexts~\citep{wu2025retrieval} and for
ranking pages in visual documents~\citep{li2026multimodalheads}.
Region-level analyses trace information about a referred region
through a vision--language model~\citep{park2026vrh}.
\citet{park2026irh} distinguish a head's attention to an image from
the image identity recoverable from its output. We use a pointer
head as a source of candidate rankings. For our inference procedure,
the relevant quantities are how often the shortlist retains the
target and how often a grounded check recovers it when rank one
is wrong.

\paragraph{Checking evidence before answering.}
Visual retrieval pipelines adapt the number of retrieved pages or
combine evidence from multiple retrieved images
\citep{sun2025visrag2,mialland2026visar,hu2026visdocagentbench}.
HIVE verifies hypotheses to rerank multimodal search
results~\citep{abdalla2026hive}; VERA uses uncertainty to trigger
visual evidence verbalization~\citep{pei2026vera}. DoubleTake checks
source images to resolve disagreements between proposed
answers~\citep{xue2026doubletake}. In GAVR, the referring phrase
already supplies the property to verify. We query each candidate
for that property before decoding any answer, so verification can
correct the image selection without relying on an answer proposal.

\section{Methodology}
\label{sec:method}

\begin{figure*}[t]
  \centering
  \includegraphics[width=\textwidth]{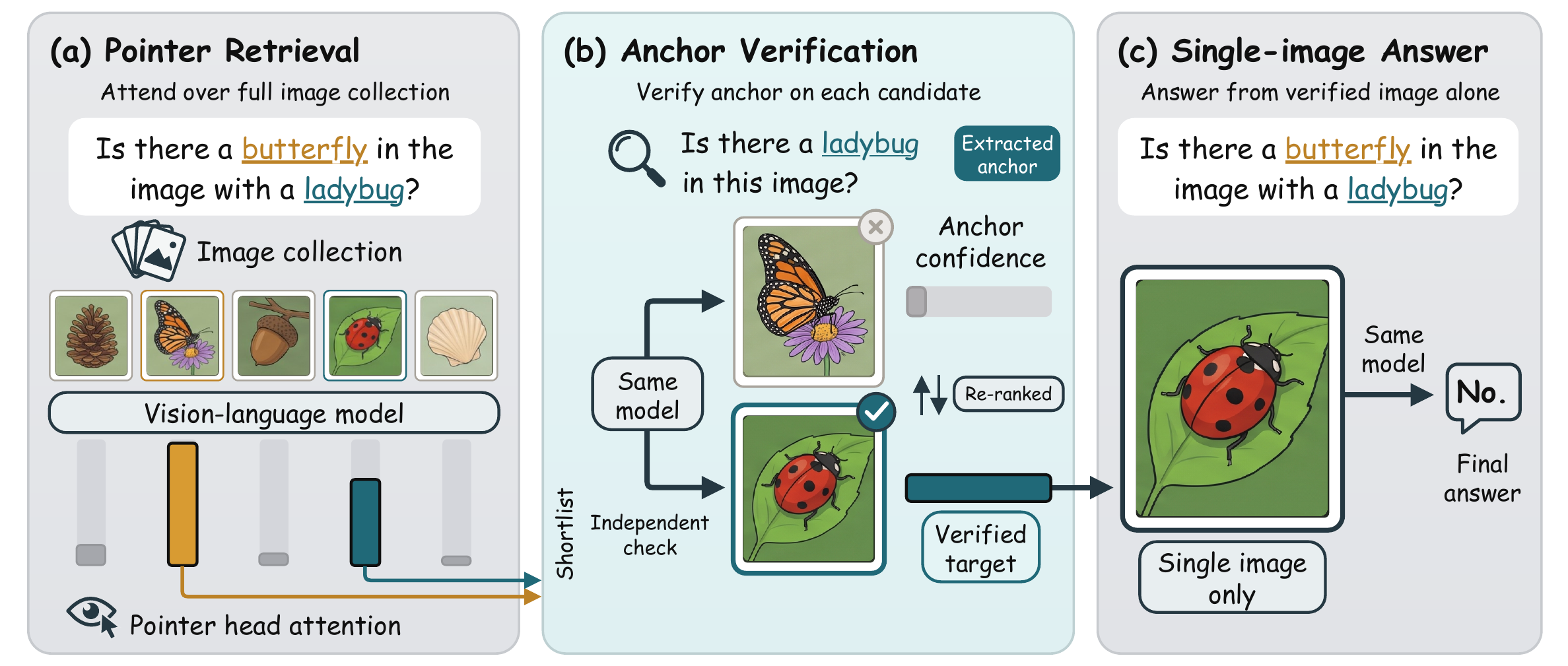}
  \caption{GAVR inference. (a) Pointer-head attention ranks the image
  collection and proposes a shortlist. (b) The frozen vision-language
  model scores each candidate for the anchor, here a ladybug.
  (c) The same model answers the original question from the selected
  image alone.}
  \label{fig:gavr-framework}
\end{figure*}

Given a question and a collection of candidate images, GAVR returns
an answer conditioned on a single selected image. As shown in
Figure~\ref{fig:gavr-framework}, a pointer score ranks the images to
form a shortlist. The frozen model checks each shortlisted image for
the referring anchor, then answers the original question using the
image with the highest verification score. We first define the
retrieval statistic and decompose answer accuracy into terms for
candidate coverage, verification error, and single-image readout.

\subsection{Problem setup and pointer ranking}

Let $x_{1:N}$ denote the image collection, $q$ the question, and
$z^*$ the index of its target image. The referring phrase in $q$
names an anchor $\tau$ (e.g., sofa), while the rest of $q$ asks about
an attribute of the referenced image (e.g., whether it contains a
bottle). The two parts of the question can therefore be asked in
sequence: first establish the reference, then query the attribute.
GAVR uses a frozen model $\pi_\theta$ for both operations.

The retrieval signal is computed from the attention weights of a
calibrated pointer head in the multi-image context $c=(q,x_{1:N})$.
Following retrieval-head work~\citep{wu2025retrieval,park2026irh},
we take the query at anchor position $p$ and aggregate attention over
the visual tokens belonging to each image:
\begin{equation}
  r_c(i)
  \;=\;
  \sum_{j\in\mathrm{span}(x_i)}
  \mathrm{softmax}\!\left(
    \frac{\langle \mathbf{q}_{p},\mathbf{k}_{j}\rangle}{\sqrt{d_h}}
  \right)_j ,
  \label{eq:pointer_statistic}
\end{equation}
Here $d_h$ is the head dimension and the softmax normalizes over
positions visible to $p$. A high score makes an image a candidate,
but the ranking offers no explicit check that the image contains the
anchor. The head can, for example, attend to the object being queried
instead of the object that identifies the image.

\subsection{Candidate retention}

We take $S_k(c)=\operatorname{top\text{-}}k(r_c)$, the indices of the
$k$ highest-scoring images, and denote its target coverage by
$C_k=\Pr[z^*\in S_k(c)]$. For $k=1$, coverage equals pointer R@1.
The sets are nested, so coverage is nondecreasing in $k$, and
$C_k-C_1$ is the fraction of examples where rank one misses the
target but the shortlist still permits recovery. The verifier
cannot repair a miss outside the shortlist. Larger shortlists can
also offer more distractors to the verifier, so increased coverage
alone does not guarantee better answer accuracy. We choose $k$ on
a calibration split and measure this trade-off in
Table~\ref{tab:k}.

\subsection{Grounded anchor verification}

The verifier asks whether one candidate contains the anchor. It
uses only that image and an anchor question $\varphi_\tau$, so the
question no longer mentions the object or attribute that the final
answer is about. We use the first-token affirmative probability
\begin{equation}
  v_i
  \;:=\;
  \pi_\theta\!\left(\langle\textsc{yes}\rangle
  \;\middle|\;
  \mathbf{V}(x_i),\;\varphi_\tau
  \right),
  \label{eq:verifier}
\end{equation}
to select $\hat z=\arg\max_{i\in S_k}v_i$. The anchor is extracted
from the referring phrase without a model call. Under the binary
interpretation of the score, $v_i$ and its log odds
$\log[v_i/(1-v_i)]$ give the same ranking. The model can thus compare
evidence for the anchor across candidates without fitting a separate
reranker or generating a free-form explanation. The check is
informative when the target's anchor is visually recognizable and
uncommon among retained distractors. If several images contain it, or
the verifier confuses a distractor with the anchor, a high score can
still lead to the wrong image. Appendix~\ref{sec:method_boundary}
describes these conditions.

Once selected, $x_{\hat z}$ is passed to the same frozen model with
the original question. We keep that question intact so the readout
answers the requested attribute, rather than the presence question
used by the verifier. With a single candidate, inference reduces to
pointer top-1 readout. For larger $k$, the pointer proposes the
candidate set and the verifier determines which image conditions
the answer.

\subsection{Selection and readout errors}

Let $C_k$ be the probability that $S_k$ contains the target and
$\beta_k$ the conditional probability of selecting a distractor given
that the target is retained. Let $\rho$ be the conditional accuracy
on successful selection, and $\rho_1$ and $\rho_0$ the accuracies in
the two failure cases described below.
\begin{lemma}[Selection--generation decomposition]
\label{lem:decomp}
For the pipeline ``retain $S_k$ $\rightarrow$ verify within $S_k$
$\rightarrow$ read out on the selected image'', the end-to-end accuracy
satisfies the exact identity
\begin{equation}
  \mathrm{Acc}(k)
  =
  C_k\Big[(1-\beta_k)\,\rho \;+\; \beta_k\,\rho_1\Big]
  \;+\;(1-C_k)\,\rho_0 ,
  \label{eq:decomp}
\end{equation}
where $\rho_1\le\rho$ is the readout accuracy on a \emph{wrong} image.
Consequently
\begin{equation}
  \mathrm{Acc}(k)
  \;\ge\;
  C_k\,(1-\beta_k)\,\rho .
  \label{eq:decomp_lb}
\end{equation}
\end{lemma}

Here $\rho_1$ conditions on selecting a distractor from a shortlist
that contains the target, while $\rho_0$ conditions on a shortlist
that misses the target. This decomposition distinguishes the effect
of increasing candidate coverage from the effect of improving the
verifier. It also distinguishes retrieval errors from errors made
while answering with the correct image. Appendix~\ref{app:proofs}
gives the proof, and Appendix~\ref{app:inference} specifies decoding
and the optional fusion rule.

\subsection{Why the number of distractors matters}

In a stylized dilution model, an isolated image gives a positive
answer margin $\Delta_1$. If each distractor contributes leakage
$\varrho\ge 0$, joint conditioning reduces that margin to
\begin{equation}
  \Delta_N \;=\; \frac{\Delta_1}{1+(N-1)\,\varrho},
  \qquad
  \mathrm{Acc}_{\mathrm{joint}}(N)
  \;=\;
  \varsigma\!\left(c\,\Delta_N\right),
  \label{eq:dilution}
\end{equation}
where $\varsigma$ is the logistic function and $c>0$ sets the
margin scale. This motivates two predictions. \textbf{(P1)} At
comparable shortlist coverage, the advantage of focused readout
over joint answering grows with the number of distractors.
\textbf{(P2)} With few images and strong initial retrieval, there
is less room for verification to improve accuracy.
Appendix~\ref{app:theory} gives the assumptions and the corresponding
analysis of focused conditioning.

\subsection{Inference procedure and cost}

The scoring pass, verification passes, and answer pass have
different roles. The pointer requires an attention statistic from
the image collection. After that statistic is available, the
verifier processes the retained images individually and releases
each candidate's visual context before evaluating the next.
The selected image is then read out in a fresh forward pass.
With a fixed candidate budget, verification and readout require
at most $k+1$ single-image passes. In the canonical setup, each
such pass uses 256 visual tokens. This accounting applies to
verification and readout; the pointer computation still depends
on the supplied collection. Appendix~\ref{app:inference} gives
the algorithm and optional margin-based fusion.

\subsection{When verification can improve on top-one readout}

Define $T_k=\Pr[\hat z=z^*]$, the probability of choosing the correct
image after ranking and verification. The selection step cannot
choose the target when it lies outside $S_k$, so
\begin{equation}
  T_k=C_k(1-\beta_k),\qquad
  T_k>C_1\iff\beta_k<1-\frac{C_1}{C_k},
  \label{eq:selection-gain}
\end{equation}
for $C_k>0$. This condition states how much verification error
can be tolerated for a given increase in coverage. If $C_k=C_1$,
the shortlist adds no recoverable targets. If $C_k$ rises
substantially, the verifier can still improve selection despite
some errors among the retained candidates.

\paragraph{Recovery and reversals.}
On the same questions, let $P$ indicate that the pointer's top-ranked
image is correct, and let $V$ indicate that GAVR selects the target.
Then
\begin{equation}
  T_k-C_1=\Pr[V=1,P=0]-\Pr[V=0,P=1].
  \label{eq:recovery-reversal}
\end{equation}
The first term counts recovered selections and requires the target
to be present below rank one. The second term counts reversals
of an initially correct choice. These events can both occur when
coverage is high, which is why the main comparison holds the final
readout fixed and changes how its image is chosen. The qualitative
examples in Figure~\ref{fig:cases} illustrate a recovery and a case
where the verifier selects a distractor.

\paragraph{From selection to answer accuracy.}
If two selection procedures share the same conditional accuracy
$\rho$ on the target and the same conditional accuracy $b$ on
a distractor, Eq.~\eqref{eq:decomp} reduces to
\begin{equation}
  \mathrm{Acc}(k)-\mathrm{Acc}(1)=(\rho-b)(T_k-C_1).
  \label{eq:matched-readout}
\end{equation}
This special case isolates the contribution of selecting the right
image. In practice, the distractors selected by the two procedures
can differ and still produce the correct answer, for example when
the queried object is absent from both. Selection accuracy and
answer accuracy should therefore be reported separately. The
selection-aware error analysis in Section~\ref{sec:exp_attribution}
uses cases in which the target is absent, cases where verification
misses a retained target, and correct selections followed by an
incorrect answer.

\paragraph{Confidence after selection.}
The model generates the final answer from the original question
and the verified image. Its first-step margin compares the two
most likely answer tokens before decoding further. A small margin
can indicate ambiguity in the selected image or uncertainty about
the queried attribute. It measures confidence in the readout and
does not directly measure whether the selected image is the target.
We bucket examples by this margin to test how well the signal
tracks accuracy in practice. The reported GAVR comparisons use
greedy readout, while Appendix~\ref{app:inference} describes the
optional rule that also considers the joint model's margin.

\paragraph{Stage-level evaluation.}
For each example, we retain the original pointer ranking, the
shortlist, the verification scores, the selected image, and the
answer. The target index is used only for evaluation. Our analysis
stratifies examples by whether the target survives retrieval and
whether verification chooses it. This yields disjoint error groups:
a miss outside the shortlist, a wrong choice within a shortlist
that contains the target, and an incorrect answer on the target
image. Comparing the groups at fixed readout resolution allows us
to attribute changes to the stage where they are observed.

\paragraph{Matched comparisons.}
To measure what verification changes, we compare GAVR with
pointer top-1 on the same questions, using the same answer prompt
and image resolution. The comparison with joint inference instead
changes both image selection and the context for generation.
An improvement over joint inference alone cannot isolate the
gain from recovering a target below rank one.

\clearpage
\section{Experiments}
\label{sec:experiments}

\paragraph{Evaluation.}
We evaluate Visual Haystacks~\citep{wu2025visual} at
$N\in\{2,3,5,10,20,50\}$ with $n{=}1000$ paired questions per
setting. ICIR-MCQ has 6{,}000 questions with five or six images;
accuracy averages four option rotations. We use five frozen models
(InternVL3.5-4B/8B/14B and Qwen3-VL-4B/8B), $k{=}5$, and greedy
decoding at $448^2$ resolution. Base Joint answers from all images;
Pointer top-1 answers from the highest-ranked image. Head choices,
controls, paired tests, and replication are detailed in
Appendix~\ref{sec:exp_protocol}, including the controls and
replication protocols.

\subsection{Haystack scaling}
\label{sec:exp_scaling}

Table~\ref{tab:main-vhs} evaluates the central prediction (P1).
Figure~\ref{fig:scaling} shows how the gap changes with $N$. Joint
answering does not merely degrade: it collapses as distractors
accumulate.  On InternVL3.5-8B the joint baseline falls from 0.843 at
$N{=}2$ to 0.535 at $N{=}50$, a loss of 30.8 points; GAVR loses 14.1
points over the same range and holds 0.740.  The asymmetry appears on
every model: the joint decline is 22.2--30.8 points, the GAVR decline
11.8--14.1 points.  Two readings follow from the dilution model of
\S\ref{sec:method_frc}.  First, part of GAVR's own decline is coverage,
not readout: at $N{=}50$ the pointer's R@1 drops further and some
targets leave the top-5, which no verification can undo.  Second, the
readout channel is protected: the same frozen model answers the same
single-image question at the same resolution regardless of $N$, so the
only $N$-dependence left in GAVR is the candidate pool, an explicitly
audited quantity.

\begingroup
\baselineskip=\dimexpr\baselineskip\relax plus .25pt
elax
All 24 (model, $N$) cells improve over Base Joint, 22 of them by at
least five points.  The gain grows from $+3.8$ to $+9.3$ points at
$N{=}2$, through $+16.1$ to $+18.8$ points at $N{=}10$, and reaches
$+18.4$ to $+20.5$ points at $N{=}50$, flattening only when the target
coverage itself saturates.  The four $N{=}50$ gains are statistically
decisive: McNemar $p=1.8\times10^{-19}$ (InternVL3.5-8B),
$2.3\times10^{-17}$ (InternVL3.5-4B), $6.6\times10^{-24}$ (Qwen3-VL-8B),
and $1.4\times10^{-30}$ (Qwen3-VL-4B), with all 95\% bootstrap CIs in
$[+14.4,+24.8]$ points.  No model at any $N$ is worse with GAVR.

The absolute numbers give the result its practical shape: GAVR at
\emph{fifty} images matches or beats the joint baseline at \emph{five}
images on every model (0.740--0.770 vs.\ 0.736--0.756), and it exceeds
the joint baseline at ten images by 7.3--9.4 points.  An application
that can tolerate five images under joint decoding can therefore
tolerate fifty under GAVR at the same accuracy.

\begin{table*}[t]
\caption{Visual Haystacks accuracy ($n{=}1000$ per cell) as a function
of haystack size $N$.  Each cell reports Base Joint / \textbf{GAVR}
($k{=}5$; every GAVR entry is bold and every entry is an improvement).
The joint baseline collapses monotonically; GAVR's decline is roughly
half as steep and saturates at the coverage of the top-5 set.}
\label{tab:main-vhs}
\centering
\setlength{\tabcolsep}{10pt}
\begin{tabular}{l cccc}
\toprule
 & InternVL3.5-8B & InternVL3.5-4B & Qwen3-VL-8B & Qwen3-VL-4B \\
$N$ & Base / \textbf{GAVR} & Base / \textbf{GAVR} & Base / \textbf{GAVR} & Base / \textbf{GAVR} \\
\midrule
2  & 0.843 / \textbf{0.881} & 0.850 / \textbf{0.886} & 0.804 / \textbf{0.897} & 0.821 / \textbf{0.884} \\
3  & 0.797 / \textbf{0.868} & 0.821 / \textbf{0.875} & 0.780 / \textbf{0.895} & 0.780 / \textbf{0.880} \\
5  & 0.739 / \textbf{0.856} & 0.746 / \textbf{0.871} & 0.756 / \textbf{0.885} & 0.736 / \textbf{0.871} \\
10 & 0.654 / \textbf{0.834} & 0.681 / \textbf{0.842} & 0.678 / \textbf{0.866} & 0.672 / \textbf{0.856} \\
20 & 0.603 / \textbf{0.798} & 0.611 / \textbf{0.798} & 0.638 / \textbf{0.841} & 0.603 / \textbf{0.831} \\
50 & 0.535 / \textbf{0.740} & 0.570 / \textbf{0.754} & 0.582 / \textbf{0.770} & 0.562 / \textbf{0.766} \\
\bottomrule
\end{tabular}
\end{table*}

\begin{figure*}[t]
  \centering
  \includegraphics[width=\textwidth]{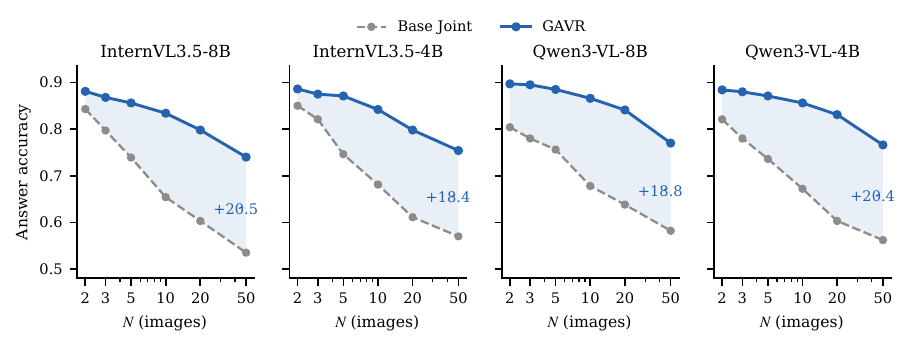}
  \caption{Accuracy versus haystack size (recorded full-scale runs,
  $n{=}1000$ per point; same data as Table~\ref{tab:main-vhs}).  The
  shaded band is the gap recovered by verification and re-conditioning;
  the annotation gives the $N{=}50$ gain in points.  The joint baseline
  falls steeply on every model; GAVR degrades with the coverage of the
  top-5 set rather than with the readout.}
  \label{fig:scaling}
\end{figure*}

\begin{figure*}[t]
  \centering
  \includegraphics[width=\textwidth]{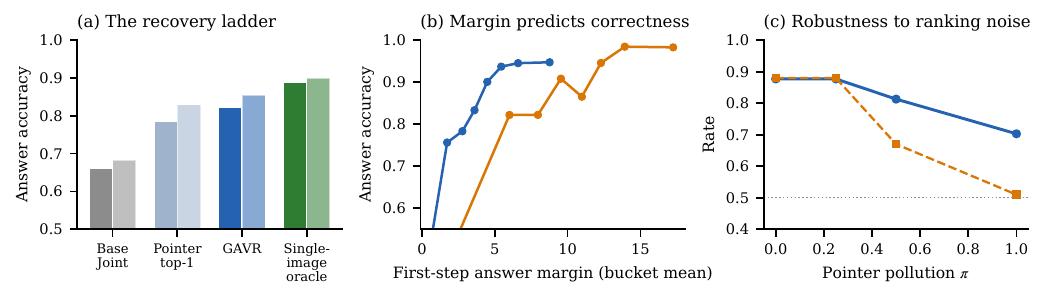}
  \caption{Factor attribution.  (a) The recovery ladder on the
  replication runs ($n{=}1000$): within each method group the dark bar
  is InternVL3.5-8B and the light bar Qwen3-VL-8B; verification closes
  over a third of the top-1-to-oracle gap.  (b) Answer accuracy by
  first-step margin bucket: blue is InternVL3.5-8B, orange
  Qwen3-VL-8B; near-monotone on both models, spanning 0.518--0.947 and
  0.534--0.983.  (c) Accuracy and verified-selection rate as the
  pointer is corrupted with probability $\pi$ (InternVL3.5-4B): solid
  curve is accuracy, dashed curve is verified selection; graceful
  degradation, and the verifier keeps a substantial signal even when
  the pointer is pure noise.}
  \label{fig:attribution}
\end{figure*}

\subsection{The commit-and-answer baseline}
\label{sec:exp_top1}

Prediction (P3) concerns the strongest baseline in the literature's own
terms: the pointer head used the way retrieval heads are meant to be
used---take its top-1 image and read the answer from it.  This pipeline
is the closest prior position to GAVR, so it is the comparison that
matters.  Table~\ref{tab:top1} and Figure~\ref{fig:top1} show that it
loses to GAVR on all five models, by 1.5 to 6.6 points, with a mean gain
of 5.1 points.

Three observations sharpen the result.  First, the ranking signal
behind the top-1 column reproduces the published single-head R@1 range
(0.787--0.836), so the comparison is not against a strawman but against
the signal at its strongest.  Second, the ordering of the gains matches
the ordering of the retrieval quality: the largest gap, +6.6 on
Qwen3-VL-8B, and +5.5 on InternVL3.5-14B, occur where the raw head is
weakest relative to the readout, and the smallest, +1.5 on
InternVL3.5-4B, occurs where R@1 is already 0.827 and the joint
baseline is strong.  Third, the gap is structural rather than
parametric: no commit-and-answer pipeline, however good its ranking,
can recover a target it has discarded, so its accuracy is capped at
$C_1\rho+(1-C_1)\rho_1$ by Proposition~\ref{prop:ceiling}, while GAVR
converts the same signal into a $C_5$-coverage problem with a second,
decorrelated adjudicator.

The replication stack runs the same comparison with a re-derived
single-image decoder and reproduces the effect: top-1 0.785 vs.\ GAVR
0.823 on InternVL3.5-8B (+3.8) and 0.829 vs.\ 0.855 on Qwen3-VL-8B
(+2.6) at $N{=}10$, $n{=}1000$.  The direction of the gap is
stack-invariant; its size shifts by 0.9 and 4.0 points across stacks,
reflecting how sensitive a single greedy decode is to parser and stack
details.

\endgroup

\begin{table}[t]
\caption{Pointer top-1 followed by single-image answering vs.\ GAVR at
$N{=}10$, $n{=}1000$, identical $448^2$ protocol, identical pointer
head per model.  The pointer heads match the published single-head
retrieval; GAVR wins on every model, most where retrieval is weakest.}
\label{tab:top1}
\centering
\small
\setlength{\tabcolsep}{5pt}
\begin{tabular}{l ccc}
\toprule
Model & Pointer top-1 & \textbf{GAVR} & $\Delta$ \\
\midrule
InternVL3.5-8B  & 0.787 & \textbf{0.834} & $+4.7$ \\
InternVL3.5-4B  & 0.827 & \textbf{0.842} & $+1.5$ \\
InternVL3.5-14B & 0.780 & \textbf{0.835} & $+5.5$ \\
Qwen3-VL-8B     & 0.800 & \textbf{0.866} & $+6.6$ \\
Qwen3-VL-4B     & 0.813 & \textbf{0.856} & $+4.3$ \\
\midrule
Mean            & 0.801 & \textbf{0.847} & $+5.1$ \\
\bottomrule
\end{tabular}
\end{table}

\begin{figure*}[t]
  \centering
  \includegraphics[width=0.878\textwidth]{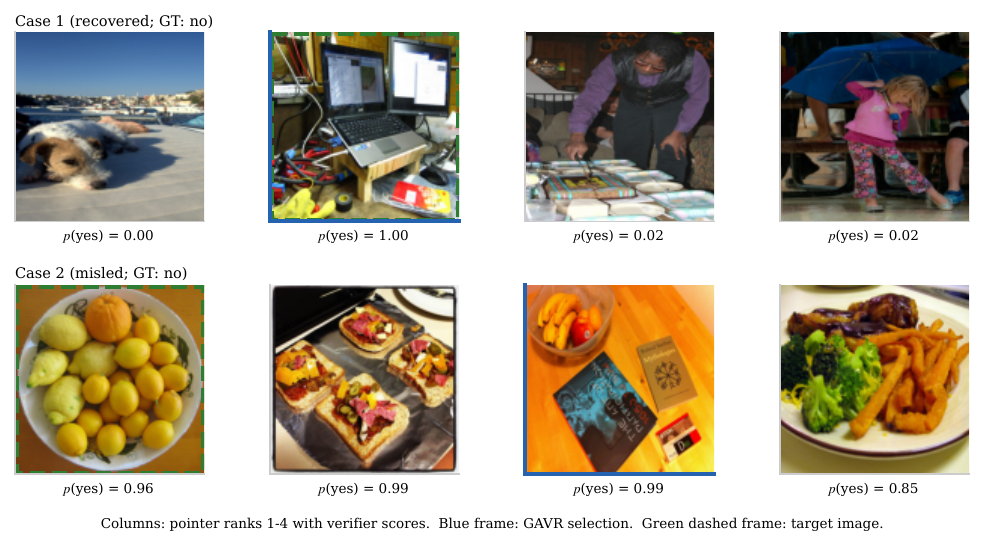}
  \caption{Two benchmark examples (InternVL3.5-8B, $N{=}10$), showing
  pointer ranks 1--4 and the verifier score under each image.  Blue
  frame: GAVR selection; green dashed frame: target.  Case 1, ``For the
  image with a mouse, is there a backpack?'': the pointer is
  query-biased (it ranks a backpack image first) while the verifier
  isolates the anchor image ($p{=}1.00$) and recovers the correct
  answer.  Case 2, ``For the image with an orange, is there a dining
  table?'': the anchor is not discriminative---the target scores 0.96
  but two distractors score 0.99---and verification is dragged to a
  wrong image; the joint baseline was right.  The pair marks the
  operating range of the method.}
  \label{fig:cases}
\end{figure*}

\begin{figure}[t]
  \centering
  \includegraphics[width=\columnwidth]{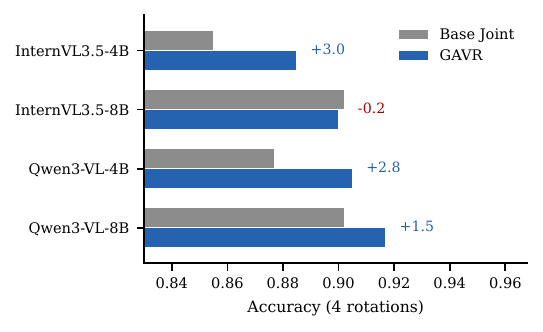}
  \caption{MCQ accuracy (four rotations, 6{,}000 examples).  Gains are
  small precisely where pointer R@1 is near ceiling; the
  InternVL3.5-8B $-0.2$ cell is within paired noise (replication:
  $+0.7$).  Compare the $N{=}50$ gains of 18.4--20.5 points in
  Table~\ref{tab:main-vhs}.}
  \label{fig:mcq}
\end{figure}

\subsection{Attribution and ablations}
\label{sec:exp_attribution}

Lemma~\ref{lem:decomp} makes the method auditable: each factor can be
measured separately, and the product must reproduce the end-to-end
number.  Figure~\ref{fig:attribution}(a) draws the recovery ladder for
the two 8B models on the replication runs: Base Joint 0.661 / 0.682,
pointer top-1 0.785 / 0.829, GAVR 0.823 / 0.855, and the single-image
oracle 0.888 / 0.900.  Verification recovers about 37\% of the
top-1-to-oracle distance and 71--79\% of the base-to-oracle gap---the
part the decomposition attributes to selection.  Averaged over the five
models, coverage is
$C_5\approx0.95$ (top-5 recall 0.950--0.956), the verifier re-ranks the
retained set to a selection R@1 of 0.86--0.91 (up from 0.79--0.83), the
oracle reads 0.901, and GAVR lands at 0.834--0.866: 94.6\% of the
oracle's answer accuracy.  The residual error budget splits cleanly into
coverage misses (about 5\% of the mass), residual verifier errors, and
a downstream generation error of 6.9\% that no selection method can
remove. Figure~\ref{fig:errbudget} repeats the decomposition on the
two 8B replication runs ($n{=}1000$ per model) and shows the same
shape---readout error is the largest single loss (8.8--10.0\%), while
coverage misses stay below 3\%.

The recovery statistic makes the mechanism concrete: of 215
InternVL3.5-8B examples whose top-1 selection was wrong, GAVR answers
136 correctly, a 63.3\% recovery rate.  These are questions that
neither the joint baseline nor the commit-and-answer pipeline can save,
because the answers they could give were measured on the wrong image.

\paragraph{Verifier calibration.}
If verification is a likelihood-ratio test, its confidence should
predict answer accuracy monotonically.  Figure~\ref{fig:attribution}(b)
buckets every example of the two 8B replication runs by the magnitude
of its first-step answer margin and plots accuracy per bucket: accuracy
climbs from 0.518 in the lowest bucket to 0.947 (InternVL3.5-8B) and
from 0.534 to 0.983 (Qwen3-VL-8B), a 43--45 point span with a single
4-point inversion mid-curve on Qwen3-VL-8B.  The margin is therefore a
usable abstention signal---a deployment can route low-margin cases to a
human or to a wider candidate set without retraining anything.

\paragraph{Candidate budget.}
Table~\ref{tab:k} sweeps $k$.  The $k{=}1$ column is the verified redraw
of the raw top-1 pick and coincides numerically with the commit-and-
answer baseline, so the sweep starts exactly where the prior pipeline
ends.  $k{=}3$ captures most of the available gain; the operating point
$k{=}5$, measured on the two models where the elbow is flattest, adds
at most 0.7 further.  Beyond the elbow, extra candidates add only
verifier false-positive surface, which is why we stop at five and why
the gain is attributable to verification rather than to candidate
count.

\begin{table}[t]
\caption{Candidate budget ablation at $N{=}10$ ($n{=}1000$).  A dash
marks a cell that was not run: at $k{=}3$ the elbow was already
reached, so only $k{=}3$ is reported for that model.  $k{=}1$
coincides with the pointer top-1 baseline of Table~\ref{tab:top1}; the
retention curve elbows at $k{=}3$--$5$.}
\label{tab:k}
\centering
\small
\setlength{\tabcolsep}{5pt}
\begin{tabular}{l ccc}
\toprule
Model & $k{=}1$ & $k{=}3$ & $k{=}5$ \\
\midrule
InternVL3.5-8B  & 0.787 & \textbf{0.837} & --- \\
InternVL3.5-4B  & 0.827 & \textbf{0.873} & \textbf{0.877} \\
InternVL3.5-14B & 0.780 & \textbf{0.857} & --- \\
Qwen3-VL-8B     & 0.800 & \textbf{0.833} & \textbf{0.840} \\
Qwen3-VL-4B     & 0.813 & \textbf{0.830} & --- \\
\bottomrule
\end{tabular}
\end{table}

\paragraph{Robustness to ranking noise.}
We corrupt the pointer statistic directly: each candidate outside the
target is promoted into the top-1 slot with probability $\pi$ before
GAVR runs (InternVL3.5-4B, $N{=}10$; Figure~\ref{fig:attribution}(c)).
At $\pi{=}0.25$ accuracy is unchanged at 0.877; at $\pi{=}0.5$ it is
0.813 with a verified selection rate of 0.670; even at $\pi{=}1.0$,
where the top-1 signal is pure noise and verified selection collapses to
0.510, accuracy is 0.703---above what text-only prompting achieves and
far above chance.  The two curves separate slowly at first: the
verifier repairs isolated retrieval errors without needing the pointer
to be right, and degrades gracefully under systemic corruption.  This is
what an independent evidence channel should look like, and it is
Corollary~\ref{cor:orthogonal} in measured form: the verifier's error
rate moves with anchor discriminability, not with the pointer's noise.

Table~\ref{tab:controls} compares the main controls. Appendix~\ref{sec:exp_controls} describes the protocols and reports additional results.

\begin{table}[t]
\caption{Controls at $N{=}10$ ($n{=}1000$).  Same contexts, prompt, and
parser throughout; the model is InternVL3.5-8B unless marked.  Controls
that emulate the prior pipelines' intuition (order robustness, answer
aggregation) stay below the joint-verification pipeline by a wide
margin.}
\label{tab:controls}
\centering
\small
\setlength{\tabcolsep}{4pt}
\begin{tabular}{l c}
\toprule
Method & Accuracy \\
\midrule
Text-only (Qwen3-VL-8B / Qwen3-VL-4B) & 0.455 / 0.465 \\
Order-permuted (InternVL3.5-4B) & 0.707 \\
Order-permuted (Qwen3-VL-8B) & 0.660 \\
Order-permuted (Qwen3-VL-4B) & 0.690 \\
First image only & 0.496 \\
Last image only & 0.470 \\
Top-5 majority vote & 0.512 \\
Image self-consistency & 0.575 \\
\midrule
Base Joint & 0.654 \\
Pointer top-1 & 0.787 \\
\textbf{GAVR} & \textbf{0.834} \\
\bottomrule
\end{tabular}
\end{table}

\subsection{Boundary condition: the ICIR-MCQ probe}
\label{sec:exp_mcq}

Prediction (P2) is tested where the dilution model says verification
must stop paying: five or six candidates, joint answering barely
diluted, retrieval already near ceiling.  Figure~\ref{fig:mcq} shows the resulting accuracy gap.  With pointer R@1 at
0.867--0.922 there is little selection uncertainty left to monetize,
and the gains shrink to $+1.5$--$+3.0$ points on three models, while
InternVL3.5-8B moves by $-0.2$ points on the recorded matrix---inside
the paired noise.  The replication stack measures the same cell at
$+0.7$ points (0.897 $\to$ 0.904) and Qwen3-VL-8B at $+0.9$ (0.908
$\to$ 0.917); the sign flip on InternVL3.5-8B across stacks is itself
the cleanest evidence that this cell is a wash rather than a gain or a
loss.  The replication's pointer configuration is also sharper (R@1
0.989 and 0.993), which strengthens rather than weakens the boundary
reading: even near-perfect retrieval buys at most 0.9 points on a
five-image set.  This is the honest edge of the method:
GAVR is not a universal free lunch; it converts \emph{uncertainty} into
accuracy, and where there is no uncertainty there is nothing to convert.

These small gains are consistent with Eq.~\eqref{eq:dilution}:
joint answering faces fewer distractors, and the initial rankings
already recover most targets.

\paragraph{Head calibration.}
The VHS-calibrated head of InternVL3.5-14B reaches only 0.677 R@1
on MCQ, and GAVR trails Base Joint by 3.7 points. This case illustrates
the need to calibrate the retrieval head for the target task
(Appendix~\ref{sec:exp_protocol}).

\subsection{Qualitative results and inference cost}

In Figure~\ref{fig:cases}, checking for a mouse recovers the
second-ranked image. In the orange case, distractors outscore
the retained target, so verification picks the wrong image.
Appendix~\ref{sec:exp_qual} gives both readouts.

At $N{=}50$, joint prefill uses 12{,}800 visual tokens and exceeds
the memory of a 48\,GB A40. GAVR completes the same grid using
256 visual tokens per verification or readout pass. Appendix~\ref{sec:exp_cost}
reports end-to-end timing and token counts.

\section{Conclusion}

GAVR retains several images, checks each for the question's anchor,
and answers from the image with the strongest verification score.
This separates candidate coverage, verification errors, and readout
accuracy. On Visual Haystacks, GAVR gains \GainAvg{} points over
pointer top-1 at $N{=}10$ and \GainNfifty{} points over joint inference
at $N{=}50$. The smaller gains on ICIR-MCQ and the failure cases
define where this separation helps: the correct image must survive
retrieval, and its anchor must distinguish it from competing images.
Low first-step answer margins flag examples on which the
model is less likely to answer correctly, offering a practical
signal for abstention. With discriminative anchors, verification
can recover a lower-ranked target while keeping distractors
out of the answer context.

\clearpage
\bibliographystyle{icml2027}
\bibliography{sample,gavr-method-refs}

\clearpage
\appendix
\input{appendix}
\end{document}

%% file: appendix.tex
\section{Methodology}

\subsection{Extended analysis of GAVR}
\label{app:theory}

\subsubsection{Problem setting and notation}
\label{sec:method_setup}

Let $\pi_\theta$ be a frozen large vision--language model (LVLM) that
consumes an interleaved context and emits text.  Each image $x_i$ is
encoded into a fixed-length block of $t$ visual tokens
$\mathbf{V}(x_i)\in\mathbb{R}^{t\times d}$.  A \emph{visual-haystack}
(problem instance)
\[
  c \;=\; \big(q,\; x_1,\dots,x_N\big)
\]
consists of a yes/no question $q$ and $N$ candidate images, exactly one of
which---the \emph{target}---is the image the question refers to.
We write $z^\ast\in\{1,\dots,N\}$ for the identity of the target and
$a^\ast\in\{\textsc{yes},\textsc{no}\}$ for the gold answer.  In the
benchmark family we consider, $q$ mentions an \emph{anchor object} $\tau$
through a grounding phrase (e.g., \emph{``in the image with a sofa, is
there a bottle?''}, where $\tau=\textsf{sofa}$ is the anchor and
\textsf{bottle} is the queried object), and $\tau$ is visually
present \emph{only} in $x_{z^\ast}$.  This anchor mechanism is what makes
the selection problem well-posed; we make the corresponding assumption
explicit in \S\ref{sec:method_boundary}.

A model that answers $q$ over $c$ must implicitly solve two subproblems:
\emph{selection} (which image is $x_{z^\ast}$?) and \emph{readout}
(given that image, is there a bottle?).  The classical
\emph{joint} predictor commits to both at once,
\begin{equation}
  \hat a_{\mathrm{joint}}(c)
  \;=\;
  \arg\max_{a}\;
  \pi_\theta\!\left(a\mid q,\mathbf{V}(x_1),\dots,\mathbf{V}(x_N)\right),
  \label{eq:joint_predictor}
\end{equation}
i.e., selection is latent and never externalised or checked.
Every component of GAVR leaves $\pi_\theta$ untouched: it acts only on
\emph{which conditioning inputs} the frozen model is queried with, and in
\emph{which order}.

The pointer score is defined in Eq.~\eqref{eq:pointer_statistic}.

\subsubsection{A selection--generation decomposition}
\label{sec:method_decomposition}

Write $S_k(c)\subseteq\{1,\dots,N\}$ for the top-$k$ images under
$r_c$, and define the \emph{coverage} of the candidate set,
\begin{equation}
  C_k \;=\; \Pr\!\left[\, z^\ast \in S_k(c) \,\right],
  \qquad
  C_1 \;=\; \text{retrieval R@1 of } r_c .
  \label{eq:coverage}
\end{equation}
Conditioned on coverage, let
$\beta_k=\Pr[\text{verifier discards }z^\ast\mid z^\ast\in S_k]$ be the
verifier's error rate \emph{within the candidate set},
$\rho$ the probability that the frozen model answers correctly when it is
conditioned on the true image alone, and $\rho_0$ the probability that it
answers correctly without the target in its candidate set.
Lemma~\ref{lem:decomp} in the main text states the accuracy identity.

Three properties of \eqref{eq:decomp} organise everything that follows.
\begin{enumerate}
\item \textbf{Each mechanism controls one factor.}
$C_k$ is a property of the candidate generator, $\beta_k$ of the
verifier, and $\rho$ of the readout conditioning.  Hence the design
problem splits into three subproblems that can be solved independently
and composed without interference---this is the formal justification for
GAVR being a \emph{stack} of three components rather than one monolithic
prompt.
\item \textbf{Joint answering is the degenerate corner.}
The joint predictor \eqref{eq:joint_predictor} corresponds to $k=1$,
$\beta_1=0$ and a readout $\rho$ measured \emph{jointly} over all $N$
images.  Its accuracy is therefore bounded by the single-head R@1 times
the readout quality, with \emph{no recovery path} for selection errors.
\item \textbf{Uncertainty is where the gain is.}
Comparing \eqref{eq:joint_predictor} with \eqref{eq:decomp} isolates the
\emph{selection--generation gap} $\mathrm{Acc}_{\text{oracle single}} -
\mathrm{Acc}_{\text{joint}}$, where the oracle uses $S=\{z^\ast\}$.  On
our benchmarks this gap reaches ${\sim}25$pp at $N{=}10$
($\AccOracle$ vs.\ the joint baseline of \S\ref{sec:experiments}), and
\RecoveredPct{} of joint retrieval failures are recoverable once the true
image stays inside the candidate set.  GAVR is designed to recover
exactly this gap, and nothing more: it cannot fix readout errors
($\rho<1$) and it cannot fix images that left $S_k$.
\end{enumerate}

\begin{corollary}[Orthogonal composition]
\label{cor:orthogonal}
Suppose mechanism $A$ increases $C_k$ at fixed $\beta_k,\rho$, mechanism
$B$ decreases $\beta_k$ at fixed $C_k,\rho$, and mechanism $C$ increases
$\rho$ at fixed $C_k,\beta_k$.  Then each mechanism can only increase the
right-hand side of \eqref{eq:decomp}, and their composition gains are
additive in the log-odds of the per-sample success probabilities.  In
particular, no mechanism can \emph{cancel} the effect of another short of
driving one factor to a regime where the other factors are irrelevant
(e.g., $C_k\to1$).
\end{corollary}

Proofs of the decomposition and the propositions below are collected in Appendix~\ref{app:proofs}.

\subsubsection{Candidate retention}
\label{sec:method_hse}

The target can rank below a distractor when the pointer score is noisy.
For a fixed context, $r_c(i)$ in \eqref{eq:pointer_statistic} combines
a signal about the target posterior with head-specific distortion and
noise from competition among the $N$ images. We model this score as
\begin{equation}
  r_c(i) \;=\; g\!\left(p_i\right) \;+\; \varepsilon_i ,
  \qquad
  p_i := \Pr\!\left[z = i \mid c\right],
  \label{eq:noise_model}
\end{equation}
where $g$ is a strictly increasing link and $\varepsilon_i$ are
zero-mean perturbations with typical scale $\sigma$ that may be
\emph{correlated across images} (the softmax couples them: raising one
image's mass lowers the others').  Under \eqref{eq:noise_model}, the
argmax is the Bayes rule only when $\sigma\to 0$; with $\sigma>0$ the
top-1 decision has strictly higher error than the Bayes error whenever the
true posterior is not dominated by a single atom at distance $\gg\sigma$.
Empirically, the single head attains R@1 of $\RoneIV$ (InternVL3.5-8B) and
$\RoneQV$ (Qwen3-VL-8B) at $N{=}10$, i.e., \emph{one in five questions is
answered on the wrong image before generation even starts}.

GAVR retains $S_k=\mathrm{top}\text{-}k\big(r_c\big)$ so that the
verifier can consider images ranked below the first
(\S\ref{sec:method_gav}). The following proposition relates the
value of retaining more images to target coverage and readout accuracy.

\begin{proposition}[Top-1 ceiling and the value of retention]
\label{prop:ceiling}
Consider any pipeline that selects one image from the evidence of $r_c$
and then answers on it, with readout accuracy $\rho$ on the target and
$\rho_1\le\rho$ otherwise.  Its accuracy is at most
\begin{equation}
  \mathrm{Acc}_{\text{top-}1}
  \;\le\; C_1\,\rho + (1-C_1)\,\rho_1
  \;\le\; C_1\,\rho + (1-C_1)\,\rho_0 .
  \label{eq:ceiling}
\end{equation}
Any verifier with $\beta_k<1$ raises this ceiling to
$\mathrm{Acc}\ge C_k(1-\beta_k)\rho$, which is strictly larger whenever
$C_k>C_1$ and $\rho_1<\rho$.  The retention is therefore a
\emph{sufficient-statistics relaxation}: $S_k$ preserves every hypothesis
whose posterior mass exceeds the $k$-th order statistic of $r_c$, and the
choice among them is delegated to a more reliable, non-attentional test.
\end{proposition}

The measured retention curve closes almost all of the gap: $C_1\approx
\RoneIV$ but $C_5\approx\CtopFive$ on VHS.  The relaxation costs $k$
additional decodes (\S\ref{sec:method_algo}), is independent of $N$, and
requires no re-training; it is the cheapest of the three mechanisms and
the precondition for the second.  A practical corollary of
\eqref{eq:ceiling} is that $k$ should be chosen at the \emph{elbow} of the
retention curve: beyond the elbow, extra candidates only add verifier
false-positive surface, which enters \eqref{eq:decomp} through
$\beta_k$.

\subsubsection{Grounded anchor verification}
\label{sec:method_gav}

The shortlist can contain several plausible images. To select one,
GAVR asks the frozen model whether each candidate contains the anchor.
The affirmative-token probability in Eq.~\eqref{eq:verifier} gives the verification score, where $\varphi_\tau$ asks whether the image contains the anchor $\tau$ and requests a yes-or-no answer. The term $\pi_\theta(\langle\textsc{yes}\rangle\mid\cdot)$ is the probability
that the \emph{first} decoded token is the affirmative word, read off the
next-token distribution.  The selected image is
$\hat z=\arg\max_{i\in S_k} v_i$, and the original question is then
answered conditioned on $x_{\hat z}$ alone (\S\ref{sec:method_frc}).

\paragraph{Likelihood-ratio interpretation.}
Two structural properties distinguish \eqref{eq:verifier} from re-using
$r_c$ or from a trained retriever.

\emph{(i) It is a test statistic in the model's own likelihood class.}
The quantity $v_i$ is the model's likelihood of the hypothesis
``$x_i$ contains $\tau$''.  The log-odds $\lambda_i =
\log\frac{v_i}{1-v_i}$ is the Neyman--Pearson test statistic for this
hypothesis against its complement \emph{within the function class that
$\pi_\theta$ can express}; thresholding $\lambda_i$ is therefore the most
powerful test at any fixed level $\alpha$ in that class
\citep{neyman1933problem}.  Whatever the model knows about sofas and
bottles enters the statistic through the same weights that answered the
original question: verification is \emph{grounded} in the same concept
ontology as generation, unlike external embedding-space similarity.

\emph{(ii) It conditions on \emph{one} image and on $\tau$ only.}
Dropping the other $N-1$ images removes the $N$-dependent competition
that corrupts $r_c$ (cf.\ \eqref{eq:noise_model}); dropping the original
question removes the confound between ``is $\tau$ here'' and ``is the
bottle here''.  The test is thus deliberately \emph{narrower} than the
retrieval problem it adjudicates, which is precisely what makes its
errors decorrelated from the pointer statistic's errors.

\begin{proposition}[Verification precision under label noise]
\label{prop:verify}
Assume (i) on $S_k$, the target is the only image containing $\tau$;
(ii) the verifier has a target-set sensitivity $1-\beta$ and a
false-positive rate $\alpha$, and (iii) within $S_k$ the membership prior
is not adversarial, i.e., every candidate has prior mass at least
$\tilde C$ of being the target with $\tilde C\approx 1/k$.  Then the
probability that the re-ranked argmax is correct is
\begin{equation}
  \Pr\!\left[\hat z = z^\ast \;\middle|\; z^\ast\in S_k\right]
  =
  \frac{\tilde C\,(1-\beta)}
       {\tilde C\,(1-\beta) + (1-\tilde C)\,\alpha}.
  \label{eq:precision}
\end{equation}
In particular, verification improves on random selection within $S_k$
whenever $\alpha<(1-\beta)$, and the pipeline improves on top-1
commitment whenever
\begin{equation}
  C_k\,(1-\beta)\,\rho
  \;>\;
  C_1\,\rho + (1-C_1)\,\rho_1 ,
  \label{eq:improve}
\end{equation}
a condition that holds with margin in our benchmarks
(\eqref{eq:precision} evaluated at $\alpha\lesssim0.03$,
$\beta\lesssim.1$, $k{=}5$ reproduces the measured
$\AccGAVRIV$--$\AccOracle$ band; see \S\ref{sec:experiments}).
\end{proposition}

\begin{remark}[Anchor parsing]
In all our benchmarks, $\tau$ is recoverable from $q$ by a single
regex over the grounding phrase (the ``image with a
\emph{cola}'' pattern) and we use this rule without a language-model
call, keeping the method deterministic and the cost analysis exact.  A
model-based anchor extractor would slot into the same slot of the
pipeline; the theory only requires that the extracted phrase denote the
discriminative object of the target image (Assumption~\ref{ass:anchor}).
\end{remark}

\subsubsection{Focused re-conditioning}
\label{sec:method_frc}

After verification, GAVR uses the selected image as the sole visual
input for answering. This avoids exposing the answer generator to
distractor images that can weaken the evidence for the correct answer.
For a fixed evidence gap $\Delta_1$ (the yes/no logit margin on the
single correct image), interference among $N-1$ distractor images
attenuates the effective gap of the joint forward pass.  A stylised
first-order model that matches the qualitative behaviour of softmax
competition is Eq.~\eqref{eq:dilution}, where $\varrho\in[0,1]$ is a per-distractor leak and $\varsigma$ the
logistic function.  Equation~\eqref{eq:dilution} is a model, not a law;
its purpose is to make two predictions that the design exploits.
First, joint accuracy is \emph{monotonically decreasing} in $N$ already
at fixed selection quality---consistent with the measured drop from
${\sim}0.84$ at $N{=}2$ to ${\sim}0.54$ at $N{=}50$.  Second, the
\emph{GAVR$-$joint gap} is monotone in $N$ whenever coverage is held
fixed, because the gap between $\varsigma(c\Delta_1)$ and
$\varsigma(c\Delta_N)$ increases with $\varrho(N-1)$.  This is exactly
the pattern in the main results: ${\sim}4$pp at $N{=}2$, $\GainNfifty$pp
at $N{=}50$.  The mechanism is therefore not merely a better selector;
it also removes a systematic readout deficit that grows with context
size.

\begin{proposition}[Optimal conditioning]
\label{prop:conditioning}
Among single-image choices $i\in\{1,\dots,N\}$, the choice that
maximizes the expected reduction in answer entropy,
$\mathbb{H}(a\mid q,x_i)$, maximizes $\mathbb{I}(a\,;z=i\mid q,c)$.
Under Assumption~\ref{ass:anchor}, the score $\lambda_i$ in
\eqref{eq:verifier} is a sufficient statistic for this mutual
information within the model's likelihood class. An
$\epsilon$-optimal conditioning choice within $S_k$ is therefore
\[
  \hat z=\arg\max_{i\in S_k}\lambda_i.
\]
\end{proposition}

\paragraph{Single-image decoding.}
We decode the yes/no answer on $x_{\hat z}$ alone, using the original
question verbatim.  This is deliberate: the readout question is a
\emph{second} question to the same model under a different conditioning,
so its accuracy $\rho$ measures a genuinely different and---empirically
much smaller---error surface than $\mathrm{Acc}_{\mathrm{joint}}(N)$.
Because $\rho$ and $\beta$ live in different channels, the protocol
isolation of the appendix applies: verification must run at the
canonical $448^2$ resolution for all models, while the readout may use a
higher native resolution (the two resolutions change $\rho$ and $\beta$
separately and must not be conflated when reporting).

\subsection{Assumptions, identifiability, and the boundary of the method}
\label{sec:method_boundary}

\begin{assumption}[Anchor uniqueness]
\label{ass:anchor}
The anchor object $\tau$ appears visually only in the target image; every
distractor image is anchor-free.
\end{assumption}

\begin{assumption}[Verifier monotonicity]
\label{ass:monotone}
For the frozen model, the verifier probability $v_i$ is a monotone
function of the posterior probability that $x_i$ contains $\tau$, up to
zero-mean noise; equivalently, the score $v$ is calibrated enough that
its $\arg\max$ agrees with the posterior $\arg\max$ in expectation over
the noise.
\end{assumption}

Assumption~\ref{ass:anchor} is what the benchmark construction
guarantees; it is also exactly what fails gracefully when it is violated.
If some distractor contains $\tau$ too (a ``hard negative''), the
verifier's false-positive mass rises, $\alpha$ in
\eqref{eq:precision} increases, and the selection can be wrong even after
verification---but the retained set and the focused readout still bound
the damage: the method degrades toward the readout accuracy of an
anchor-bearing image, not toward the joint baseline's much lower
accuracy.  Assumption~\ref{ass:monotone} is the interesting one: it is
\emph{empirically testable} per model and fails on specific error
families (e.g., anchors that denote textures rather than objects, or
taxonomically adjacent false positives such as confusing a sofa with a
chair).  In that regime, the gain predicted by \eqref{eq:improve}
becomes equality and can even turn negative for small candidate pools,
where $C_1$ is already near one and $\mathrm{Acc}_{\mathrm{joint}}(N)$
is not diluted; we analyse exactly this boundary in
\S\ref{sec:experiments} on the small-candidate MCQ benchmark, where the
predicted and measured gains ($+1.5$ to $+3$pp, and at most a wash on
the strongest 8B model) agree to within sampling noise.

\paragraph{On head calibration.}
The pointer statistic is assumed given.  When the head's ranking is
miscalibrated for a new model family (the 14B model is the salient example
in our experiments, where the MCQ-optimal head differs from the
VHS-optimal one), $C_1$ degrades and with it the ceiling of
\eqref{eq:ceiling}; the remedy in practice is to calibrate the head on a
small held-out split---a $O(1)$ cost relative to inference---or to treat
the head index as a hyper-parameter and select it on the same dev split
as $k$.  We make no claim about head discovery itself; that is the object
of study of \citet{park2026irh}, and our method inherits its findings as
an input.

\paragraph{Contrast with adjacent inference-time schemes.}
\emph{Trained re-rankers} (CLIP-style similarity models) replace the
retrieval statistic with an external embedding space and require
training; they are context-free and their errors are entangled with the
reader's representation, so they cannot serve as the
``independent second opinion'' that the decomposition requires.
\emph{Self-consistency} \citep{wang2023selfconsistency} re-samples the
\emph{same} conditioning, which cannot fix a selection error committed
in that conditioning.  \emph{Critique-and-revise loops}
\citep{madaan2023selfrefine} query the model with unstructured feedback;
GAVR instead fixes the conditioning with a targeted, likelihood-ratio
test whose error rate is bounded by \eqref{eq:precision}.  \emph{Head
ablation or steering} changes the weights or attention of the model and
requires knowledge of the retrieval circuit; GAVR leaves the weights
untouched and treats the retrieval signal as a black box.  The closest
prior position is the discussion of \citet{park2026irh}, which motivates
generation-time interventions on the selection side but does not specify
or analyse one; the three mechanisms above are precisely such an
intervention, analysed at the level of its three error factors.

\subsection{Proofs}
\label{app:proofs}

\subsubsection{Selection--generation decomposition}
\begin{proof}
Partition the outcome space by the events ``$z^\ast\in S_k$'' and its
complement; on the first event, partition further by whether the verifier
selects $z^\ast$, with probabilities $1-\beta_k$ and $\beta_k$; on the
second, the answer is correct with probability $\rho_0$.  The accuracy of
each branch is $\rho$ and $\rho_1$ respectively.  All terms are
non-negative, whence \eqref{eq:decomp_lb}.
\end{proof}

\subsubsection{Top-1 ceiling and candidate retention}
\begin{proof}
Immediate from \eqref{eq:decomp} with $k=1$, $\beta_1=0$ and the same
readout model; the second inequality uses $\rho_1\le\rho_0$.
\end{proof}

\subsubsection{Verification precision}
\begin{proof}
By Bayes' rule on ``target present'' versus ``false positive'': the
target branch is correct with probability $\tilde C(1-\beta)$; each of
the other $k-1$ candidates fires spuriously with probability
$(1-\tilde C)\alpha/(k-1)$ in the symmetric case, giving the denominator;
\eqref{eq:improve} is \eqref{eq:precision} inserted into
\eqref{eq:decomp_lb} and compared with \eqref{eq:ceiling}.
\end{proof}

\subsubsection{Optimal conditioning}
\begin{proof}
By the chain rule,
$\mathbb{I}(a;z\mid q,c)=\mathbb{H}(a\mid q,c)-\mathbb{H}(a\mid z,q,c)$;
conditioning on $x_i$ replaces the mixture entropy by the component
entropy, so the expected reduction is linear in the posterior mass on
$i$.  Sufficiency follows from the Neyman--Pearson argument in
Proposition~\ref{prop:verify}: under the anchor-uniqueness assumption the
event ``$\tau$ present'' coincides with ``$z=i$'', so the likelihood
ratio is a monotone transform of the posterior odds; a maximal-likelihood
choice is posterior-optimal up to the verifier's noise floor
$\beta$.
\end{proof}

\subsection{Inference details}
\label{app:inference}

\paragraph{Optional margin-gated fusion.}
For deployments where a catastrophic verifier failure must be bounded,
GAVR can fall back to the joint answer when its logit margin dominates
the single-image margin:
\begin{equation}
  \hat a =
  \begin{cases}
    \hat a_{\mathrm{joint}}, &
    m_{\mathrm{joint}} > \kappa\, m_{\mathrm{single}},\\
    \hat a_{\mathrm{single}}, & \text{otherwise},
  \end{cases}
  \qquad \kappa\ge 1 ,
  \label{eq:fusion}
\end{equation}
where $m=\lvert\ell_{\textsc{yes}}-\ell_{\textsc{no}}\rvert$ is the
first-step margin.  Under the dilution model, $m_{\mathrm{joint}}$ is
systematically the \emph{weaker} of the two in the large-$N$ regime, so
the gate is a one-sided safety net: it can only bind when the joint
answer is unusually confident relative to the focused one, and in our
grid search over $\kappa\in[1,10]$ the optimal setting on every model is
the boundary $\kappa{=}10$, i.e., ``always trust the verified readout''.
We therefore treat fusion as an insurance mechanism and not as a
component of the method's performance.

\subsubsection{Algorithm and cost}
\label{sec:method_algo}

\begin{algorithm}[t]
\caption{GAVR inference (training-free, frozen $\pi_\theta$)}
\label{alg:gavr}
\begin{algorithmic}[1]
\REQUIRE question $q$ with anchor phrase $\tau(q)$; images $x_{1:N}$;
         pointer statistic $r_c$ (cached from prefill); budget $k$;
         fusion switch ($\kappa=\infty$ disables fusion)
\STATE $S_k \gets \mathrm{top}\text{-}k$ indices of $r_c$ \COMMENT{$O(N)$ with a max-heap}
\IF{$\lvert S_k\rvert = 1$}
  \STATE \textbf{return} $\arg\max_a \pi_\theta(a\mid q,x_{S_k[1]})$
\ENDIF
\FOR{$i\in S_k$}
  \STATE $v_i \gets \pi_\theta(\langle\textsc{yes}\rangle \mid \mathbf{V}(x_i),\varphi_{\tau(q)})$
         \COMMENT{one forward, $\le 4$ decoded tokens}
\ENDFOR
\STATE $\hat z \gets \arg\max_{i\in S_k} v_i$
\STATE $\hat a \gets \arg\max_a \pi_\theta(a\mid q, x_{\hat z})$
         \COMMENT{focused readout, optional native resolution}
\IF{fusion enabled}
  \STATE \textbf{return} $\hat a_{\mathrm{joint}}$ if
         $m_{\mathrm{joint}}>\kappa\,m_{\mathrm{single}}$ else $\hat a$
\ELSE
  \STATE \textbf{return} $\hat a$
\ENDIF
\end{algorithmic}
\end{algorithm}

Per sample, GAVR spends $k$ verification forwards of one image each and
one readout forward of one image, all with at most a handful of decoded
tokens; the prefill of the $N$-image context is shared with the
retrieval statistic and is not an extra cost of the method.  The joint
baseline, by contrast, must prefill \emph{all} $N$ visual blocks
($N\times t$ tokens) at once: at $N{=}50$ this exceeds the memory of a
48 GB device for 8B models in our setup, whereas GAVR's peak memory is
independent of $N$ because it never holds more than one full image at a
time.  The additional latency is $O(k)$ short generations with $k=5$ and
does not depend on $N$, giving large-$N$ deployments a favourable
compute profile rather than merely an accuracy gain.  All decoding is
greedy and the haystack order is fixed by a deterministic seed, so runs
are exactly reproducible.


\section{Experiments}

\subsection{Evaluation protocol and model configurations}
\label{sec:exp_protocol}

\paragraph{Benchmarks.}
\emph{Visual Haystacks} (single-needle, VHs\_large)~\citep{wu2025visual}
places one target image among $N-1$ COCO distractors and asks a yes/no
question whose referring phrase names an anchor object that is present
only in the target
(``\emph{in the image with a sofa, is there a bottle?}'').  We evaluate
$N\in\{2,3,5,10,20,50\}$ with $n{=}1000$ examples per cell.  The haystack
order is fixed by a seeded shuffle over the full JSON order, so every
method, every model, and every candidate budget are evaluated on exactly
the same $N$-tuples; per-sample outcomes are therefore pairable, which
the significance tests exploit.  \emph{ICIR-MCQ} is a 6{,}000-example
multiple-choice probe built from COCO: each question presents five or six
captioned images and four caption options, and each option serves as the
gold answer in exactly one of four rotations, which removes
option-position bias; we report accuracy averaged over the four
rotations.

\paragraph{Models and protocol.}
We freeze five vision--language models in bf16 and decode greedily: InternVL3.5 (4B, 8B, and 14B) and Qwen3-VL (4B and 8B).  Table~\ref{tab:configs} lists, per model, the pointer head
used to instantiate $r_c$ and the candidate budget.  Both families
expose the pointer circuit at the same location, layer 21 head 10, at
4B and 8B---a reproducibility observation in itself, as the two
families share no weights and no training data.  InternVL3.5-14B is the only
model whose VHS-optimal and MCQ-optimal heads differ; we handle it as a
calibration case in \S\ref{sec:exp_mcq}.  All images, including
verification inputs, are processed at the canonical $448^2$ resolution
(256 visual tokens per image).  A native-resolution readout is reported
only as a labeled ablation, because resolution moves the readout channel
$\rho$ separately from the verification channel $\beta$
(\S\ref{sec:method_frc}); that ablation lifts InternVL3.5-8B from
$0.834$ to $0.857$ at $N{=}10$ (+2.3 points) with verification held at
448, confirming that the two channels are separable and that the main
table deliberately understates the readout ceiling.  The candidate
budget is $k{=}5$ for every model and every benchmark, selected once on
dev splits at the elbow of the retention curve (Table~\ref{tab:k}) and
applied unchanged everywhere else.

\paragraph{Baselines and controls.}
\emph{Base Joint} answers from all $N$ images in one pass,
Eq.~\eqref{eq:joint_predictor}.  \emph{Pointer top-1} selects the
highest-scoring image and reads the answer from it at the same
resolution; this is the strongest pipeline any single retrieval signal
can produce, and the natural baseline for the retrieval-head line of
work.  The controls in \S\ref{sec:exp_controls} cover text-only
prompting, image-order permutation, first/last-image commitment, top-5
majority voting, and full image-consistency voting.  Every baseline and
control uses the same prompt template, the same parser, and the same
decode order as GAVR; nothing is hand-tuned per method.

\paragraph{Metrics and statistics.}
We report answer accuracy, pointer R@1, top-$k$ recall, and the
verifier's selection accuracy.  Paired correctness vectors are analyzed
with exact McNemar tests and 10{,}000-resample bootstrap 95\%
confidence intervals.  Every run writes a per-sample JSONL record
(pointer scores, candidate set, verifier scores, selection, first-step
margins, and the raw response), so every number in this section can be
recomputed from the released logs; all decoding is greedy and the run
order is deterministic.

We replicate the main comparisons in a separate software and hardware setting (Appendix~\ref{app:replication}).

\begin{table}[t]
\caption{Model configurations.  Both families expose the pointer circuit
at (layer 21, head 10) at 4B and 8B; InternVL3.5-14B is the only model
whose VHS-optimal and MCQ-optimal heads differ, and we list both.
Heads are chosen on calibration splits, never on test data.}
\label{tab:configs}
\centering
\small
\setlength{\tabcolsep}{4pt}
\begin{tabular}{l c l c}
\toprule
Model & Params & Pointer head (layer, head) & $k$ \\
\midrule
InternVL3.5-4B  & 4B  & (21, 10) & 5 \\
InternVL3.5-8B  & 8B  & (21, 10) & 5 \\
InternVL3.5-14B & 14B & (24, 0)\,/\,(26, 36)$^\dagger$ & 5 \\
Qwen3-VL-4B     & 4B  & (21, 10) & 5 \\
Qwen3-VL-8B     & 8B  & (21, 10) & 5 \\
\bottomrule
\end{tabular}

\footnotesize $^\dagger$~(24, 0) on ICIR-MCQ; (26, 36) on Visual
Haystacks.
\end{table}

\begin{figure}[t]
  \centering
  \includegraphics[width=\columnwidth]{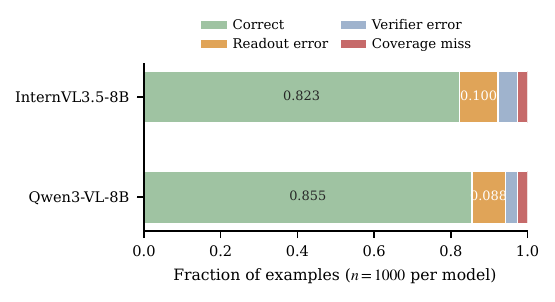}
  \caption{Error-budget decomposition on the two 8B replication runs
  ($n{=}1000$ per model): correct answers; downstream readout error
  (target selected but answered wrong); verifier error (target in the
  shortlist but another image picked); coverage miss (target never in
  the top-5).  Readout error is the largest single loss on both models;
  coverage misses stay below 3\%.}
  \label{fig:errbudget}
\end{figure}

\paragraph{Head drift on a larger model.}
InternVL3.5-14B exposes the other boundary, set by head calibration
rather than by the method.  The head that peaks on VHS (\texttt{L26H36})
has MCQ R@1 of only 0.677, and GAVR run with it trails Base Joint by
3.7 points.  A dev-split scan of the same model finds
\texttt{L24H0} / \texttt{L22H32} with R@1 of 0.843 and top-3 recall of
0.978 / 0.983---coverage restored well above the joint baseline, which
is precisely the precondition the decomposition requires
(\S\ref{sec:method_boundary}).  We report the uncalibrated number
because it marks the failure mode that \S\ref{sec:method_boundary}
predicates on a given ranking signal: the 14B MCQ cell is a
calibration case, not a method result.

\subsection{Replication protocol}
\label{app:replication}

\paragraph{Replication.}
The main matrix was produced on one software and hardware stack (four
48GB accelerators).  An independent re-implementation of the full
pipeline on a second stack (Blackwell-class GPUs, rewritten data
loader, re-derived parsers) reproduces the reported values: pointer R@1
of 0.785 and 0.829 (vs.\ 0.787 and 0.832), Base Joint of 0.661 and
0.682 (vs.\ 0.654 and 0.678), GAVR of 0.823 and 0.855 (vs.\ 0.834 and
0.866) for InternVL3.5-8B and Qwen3-VL-8B at $N{=}10$; on the
6{,}000-example probe, base 0.897 and 0.908 with GAVR 0.904 and 0.917
(vs.\ 0.902 and 0.902, GAVR 0.900 and 0.917).  All deviations are
within 1.1 points, i.e., within the paired sampling noise of the
benchmarks, and the sign and magnitude of every effect---including the
MCQ boundary---are identical across stacks.

\subsection{Controls}
\label{sec:exp_controls}

\begin{figure}[t]
  \centering
  \includegraphics[width=\columnwidth]{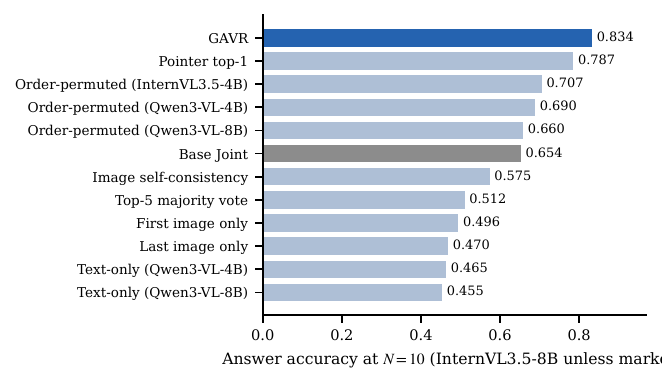}
  \caption{Every control at $N{=}10$, sorted.  The three colored bars
  are the joint baseline (gray), pointer top-1 (light blue), and GAVR
  (blue); all controls are lighter-shaded.  No control-mixture we tried
  approaches GAVR, and the two aggregation controls are the closest.}
  \label{fig:controls}
\end{figure}

Table~\ref{tab:controls} and Figure~\ref{fig:controls} rule out the
obvious alternative explanations, all measured on the same contexts,
prompt, and parser.  \emph{Seeing more images does not help}: the joint
model with permuted image order sits at 0.660--0.707, and committing to
the first or last image gives 0.496 / 0.470---i.e., the information is
in the context, and the loss is in extracting it, not in presenting it.
\emph{Aggregating answers does not help}: top-5 majority voting reaches
0.512 and full image-consistency voting 0.575, both far below GAVR
($p=2.5\times10^{-44}$ on paired samples for the latter); voting
majorities are dominated by the same distractor-attracting bias that
breaks the joint decoder.  \emph{The readout is not the problem}:
text-only prompting (with captions) reaches only 0.455--0.465, so the
gain is not an artifact of question rewriting.  And the joint model is
not a weak instantiation---it is the same 0.654 baseline that GAVR
beats at $p=1.0\times10^{-24}$.  The gain appears only when the model
is queried about the \emph{anchor} on \emph{one} image and answers on
\emph{one} image: the two steps the decomposition isolates.

Two controls deserve comment because they could, in principle, have
been strong.  Order-permutation is the standard robustness check for
positional artifacts, and it does move the joint baseline (0.660--0.707
vs.\ 0.654 for InternVL3.5-8B); the small variation across models shows that
position is a minor factor relative to dilution.  Image-consistency
voting is the best anyone can do with per-image answers alone, and its
0.575 shows that independence between answers is worth much less than
independence between the \emph{channels} that produce them.  Neither
control, nor any mixture of them we tried, approaches 0.834.

\subsection{Cost}
\label{sec:exp_cost}

\begin{figure}[t]
  \centering
  \includegraphics[width=\columnwidth]{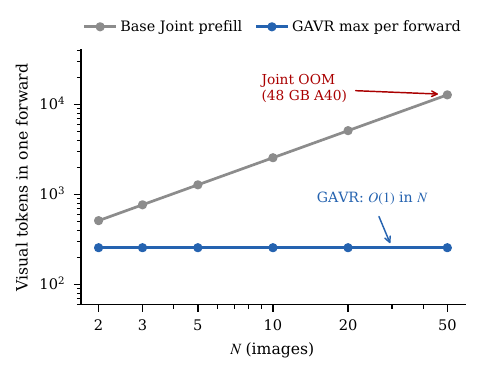}
  \caption{Visual tokens in the largest forward, as a function of
  haystack size (exact arithmetic of the protocol, not a model).  Base
  Joint grows linearly ($256N$); GAVR is flat at one image (256 tokens)
  plus text.  At $N{=}50$ the joint prefill exceeds 48GB on an A40 and
  fails, while every GAVR forward fits.}
  \label{fig:cost}
\end{figure}

GAVR adds $k$ single-image verification forwards and one single-image
readout to the prefill that already computed the pointer statistic.
With $k{=}5$ and short decodes this is a modest constant, and---the
point that matters at scale---it does not grow with $N$
(Figure~\ref{fig:cost}).  The arithmetic is exact, not asymptotic: the
joint model prefills $256N$ visual tokens in one forward, while GAVR's
largest forward holds one image plus the text context (256 visual
tokens) no matter how large the haystack grows.  Decode steps are also equal in
order: Base Joint decodes at most four answer tokens; GAVR decodes at
most four answer tokens at readout plus two per verification query.  In
return, the method adds no parameters, no gradients, and no training;
the entire inference path is deterministic under greedy decoding.

The peak-memory consequence is what separates feasible from infeasible.
On a 48GB A40, InternVL3.5-8B GAVR takes 10.57\,s per sample at $N{=}10$
end-to-end; on the replication node (97GB) the same pipeline runs at
$\approx$1.3\,s per sample, so the run-time figure is set by consumer
hardware, not by the method.  Base Joint must prefill all $N$ visual
blocks at once and at $N{=}50$ (12{,}800 visual tokens) cannot fit on a
48GB card at all, whereas GAVR completes the same $N{=}50$ grid because
its peak memory never exceeds one image plus the text context.  GAVR is
thus both more accurate and the only one of the two that scales to
large haystacks on commodity hardware---a property that follows from
the $O(1)$-in-$N$ forward in Figure~\ref{fig:cost} rather than from
optimization.

\subsection{Qualitative analysis}
\label{sec:exp_qual}

Figure~\ref{fig:cases} shows two examples from the benchmark with their
verifier scores, chosen to mark the boundary of the mechanism rather
than to flatter it.

\emph{Case 1 (recovered).}  ``For the image with a mouse, is there a
backpack?''  The pointer's first-ranked image follows the question's
queried object---a backpack---rather than the anchor, exactly the
query-bias failure mode that makes commit-and-answer pipelines
unreliable; the joint baseline answers ``yes,'' because a backpack
dominates its context.  The verifier asks ``is there a mouse?'' and the
second-ranked image scores $p(\text{yes}){=}1.00$ while every other
candidate scores at most 0.02; GAVR reads out ``no'' and is correct.
The failure of one channel is repaired by the other, which is the
mechanism Section~\ref{sec:method_frc} formalizes: the pointer's error
is diluted by the shortlist, and the verifier's evidence is
independently reliable.

\emph{Case 2 (misled).}  ``For the image with an orange, is there a
dining table?''  The target---an orange on a table---scores 0.96, but
two distractors score 0.99, and the verifier selects a wrong one; the
readout then answers ``yes'' instead of the correct ``no,'' while the
joint baseline happens to be right.  The anchor (``orange'') is
not discriminative within this shortlist, which violates anchor
uniqueness (Assumption~\ref{ass:anchor}, \S\ref{sec:method_boundary}):
when the anchor is
ambiguous, verification spends its tolerance on the wrong image.  The
two cases together delineate where the method operates: it repairs up
to the verifier's own error, and no more.  The margin distribution of
Figure~\ref{fig:attribution}(b) is the practical guard---Case 2's
near-tie (0.99 vs.\ 0.96) falls in the low-margin buckets where
accuracy drops below 0.6, i.e., flagging precisely such examples for
abstention.

\subsection{Supplementary results}
\label{app:results}

Table~\ref{tab:mcq} gives the per-model ICIR-MCQ scores behind
Figure~\ref{fig:mcq}, including pointer R@1.

\begin{table}[t]
\caption{ICIR-MCQ, 6{,}000 examples, four rotations.  Base Joint vs.\
GAVR; the pointer R@1 column shows the retrieval head operating near
ceiling, which caps any verification gain.  InternVL3.5-8B's $-0.2$ is
inside the paired confidence interval; the replication measures $+0.7$
for the same cell.}
\label{tab:mcq}
\centering
\footnotesize
\setlength{\tabcolsep}{3.5pt}
\begin{tabular}{l cccc}
\toprule
Model & Base & \textbf{GAVR} & $\Delta$ & Pointer R@1 \\
\midrule
Qwen3-VL-4B     & 0.877 & \textbf{0.905} & $+2.8$ & 0.906 \\
InternVL3.5-4B  & 0.855 & \textbf{0.885} & $+3.0$ & 0.867 \\
Qwen3-VL-8B     & 0.902 & \textbf{0.917} & $+1.5$ & 0.922 \\
InternVL3.5-8B  & 0.902 & 0.900          & $-0.2$ & 0.888 \\
\bottomrule
\end{tabular}
\end{table}

\begin{figure}[t]
  \centering
  \includegraphics[width=\columnwidth]{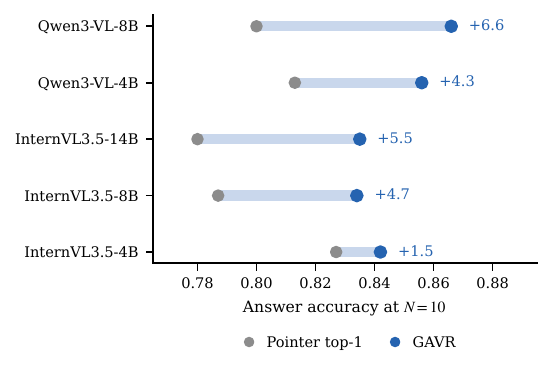}
  \caption{The commit-and-answer baseline vs.\ GAVR at $N{=}10$
  (recorded runs, $n{=}1000$).  Each dumbbell connects a model's top-1
  accuracy (gray) to its GAVR accuracy (blue); the label is the gain in
  points.  The comparison uses the same pointer head, data, prompt, and
  resolution on both ends.}
  \label{fig:top1}
\end{figure}